\pdfoutput=1

\documentclass[11pt]{article}

\usepackage[final]{acl}

\usepackage{times}
\usepackage{latexsym}

\usepackage[T1]{fontenc}

\usepackage[utf8]{inputenc}

\usepackage{microtype}

\usepackage{inconsolata}

\usepackage{graphicx}
\usepackage{algorithm}
\usepackage{algorithmic}
\usepackage{colortbl}
\usepackage{bm}
\usepackage{makecell}
\usepackage{multirow}
\usepackage{amsmath}
\usepackage{amssymb}
\usepackage{booktabs}

\title{Self-Bootstrapped Visual-Language Model for Knowledge Selection and Question
Answering}

\author{
 \textbf{Dongze Hao\textsuperscript{1,2}},
 \textbf{Qunbo Wang\textsuperscript{1}},
 \textbf{Longteng Guo\textsuperscript{1}},
 \textbf{Jie Jiang\textsuperscript{1}},
 \textbf{Jing Liu\textsuperscript{1,2}*}
\\
\\
 \textsuperscript{1}Institute of Automation, Chinese Academy of Sciences,\\
 \textsuperscript{2}School of Artificial Intelligence, University of Chinese Academy of Sciences
 \\
    \texttt{\{haodongze2021,qunbo.wang\}@ia.ac.cn},\\
    \texttt{\{longteng.guo,jie.jiang,jliu\}@nlpr.ia.ac.cn}
\thanks{Jing Liu is the corresponding author}
}

\begin{document}
\maketitle
\begin{abstract}
While large visual-language models (LVLM) have shown promising results on traditional visual question answering benchmarks, it is still challenging for them to answer complex VQA problems which requires diverse world knowledge. Motivated by the research of retrieval-augmented generation in the field of natural language processing, we use Dense Passage
 Retrieval (DPR) to retrieve related knowledge to help the model answer questions. However, DPR conduct retrieving in natural language space, which may not ensure comprehensive acquisition of image information. Thus, the retrieved knowledge is not truly conducive to helping answer the question, affecting the performance of the overall system. 
 To address this issue, we propose a novel framework that leverages the visual-language model to select the key knowledge retrieved by DPR and answer questions. 
 The framework consists of two modules: Selector and Answerer, where both are initialized by the LVLM and parameter-efficiently finetuned by self-bootstrapping: find key knowledge in the retrieved knowledge documents using the Selector, and then use them to finetune the Answerer to predict answers; obtain the pseudo-labels of key knowledge documents based on the predictions of the Answerer and weak supervision labels, and then finetune the Selector to select key knowledge; repeat. 
  Our framework significantly enhances the performance of the baseline on the challenging open-domain Knowledge-based VQA benchmark, OK-VQA, achieving a state-of-the-art accuracy of 62.83\%. Our code is publicly available at \url{https://github.com/haodongze/Self-KSel-QAns}.
\end{abstract}


\section{Introduction}
Recently, there has been an impressive advancement in large visual-language models (LVLM) \cite{li2023blip, alayrac2022flamingo, liu2023visual, Dai2023InstructBLIPTG}. They usually use a mapping network to inject visual features into the semantic space of the large language model \cite{brown2020language,zhang2022opt, touvron2023llama, vicuna, touvron2023llama} and demonstrate strong capabilities on multimodal perception and reasoning. Thus, they achieve significant progress in conventional visual question answering benchmarks \cite{antol2015vqa, goyal2017making, hudson2019gqa} which primarily focus on addressing straightforward questions that only necessitate visual perception and recognition. However, it is still challenging for the LVLMs to answer visual questions which require broader world knowledge and common sense \cite{wang2017fvqa, marino2019ok, Schwenk2022AOKVQAAB}.

Motivated by the research of retrieval-augmented generation \cite{karpukhin-etal-2020-dense} in the field of natural language processing, we use Dense Passage Retrieval (DPR) to retrieve related world knowledge to help the model answer questions. However, when using DPR, we need to transform the image into texts to retrieve the related knowledge, which leads to the underutilization of visual information. Thus, the retrieved knowledge may be unfaithful and affects the model performance. To address the issue, we consider the LVLM as the knowledge selector to find helpful knowledge from candidate retrieved knowledge by DPR. Then the selected knowledge is fed into the LVLM to predict the answer. 

In this paper, we introduce a novel framework where we adopt the large visual-language model to perform knowledge selection and question answering. 
Our framework comprises two modules: a Selector and an Answerer. We train two modules by repeating the following process: the Selector first identifies important knowledge from the candidate knowledge documents retrieved by the pre-trained retriever; then, the Answerer takes the key knowledge documents as the input knowledge and is finetuned to generate the answer; next, we generate pseudo-labels of key knowledge documents according to the Answerer's predictions and weak supervision labels; finally, we refine the Selector to assess the relevance of retrieved knowledge documents in answering the question.
This strategy of self-bootstrapping enhances the ability of knowledge selection and answer generation consistently, enabling the model to accurately respond to knowledge-intensive questions.

We conduct extensive experiments on the open-domain knowledge-based VQA benchmark (OK-VQA \cite{marino2019ok}) to validate the effectiveness of the proposed framework, where our method largely outperforms the baseline and achieves the state-of-the-art performance of 62.83\%, only finetuning 0.16\% parameters with LoRA \cite{hu2022lora}. We also conduct comprehensive ablations to validate the impact of different components of the proposed framework, including the Effect of Selector and Answerer, cycle training of the framework, varying the number of key knowledge documents, and so on.

Our contributions are summarized as follows:

\begin{itemize}
\item We introduce a novel framework that leverages the large visual-language model to select key knowledge and use them to answer questions, respectively.
\item We propose a new self-bootstrap learning method to train the Selector and Answerer, where the Selector chooses key knowledge documents for the Answerer and the Answerer provides pseudo-labels for the Selector.
\item We achieve a state-of-the-art performance of 62.83\% on the OK-VQA dataset, surpassing the previous state-of-the-art method. Notably, this improvement is achieved by fine-tuning only 0.16\% of parameters using LoRA.
\end{itemize}

\section{Related work}
\paragraph{Large Visual-Language Models.} Recently, large visual-language models \cite{li2023blip, alayrac2022flamingo, liu2023visual, Dai2023InstructBLIPTG} have demonstrated remarkable visual-language understanding and reasoning capabilities, owing to the advancement of larger language models \cite{brown2020language,zhang2022opt, touvron2023llama, vicuna, touvron2023llama}. These methods typically consist of a frozen visual encoder \cite{radford2021learning}, a visual-language connector \cite{li2023blip}, and a large language model \cite{chung2022scaling, zhang2022opt, vicuna}. The models are firstly pre-trained on large-scale visual-text datasets to align visual features to the language embedding space. After pretraining, the large language model can understand the visual details. Then, the model is finetuned to adapt to various visual-language tasks. 
In this study, we adopt BLIP2, one of the widely used models, as our backbone for bootstrapping knowledge selection and question answering with it.


\begin{figure*}[t]
  \centering
   \includegraphics[width=1.0\linewidth]{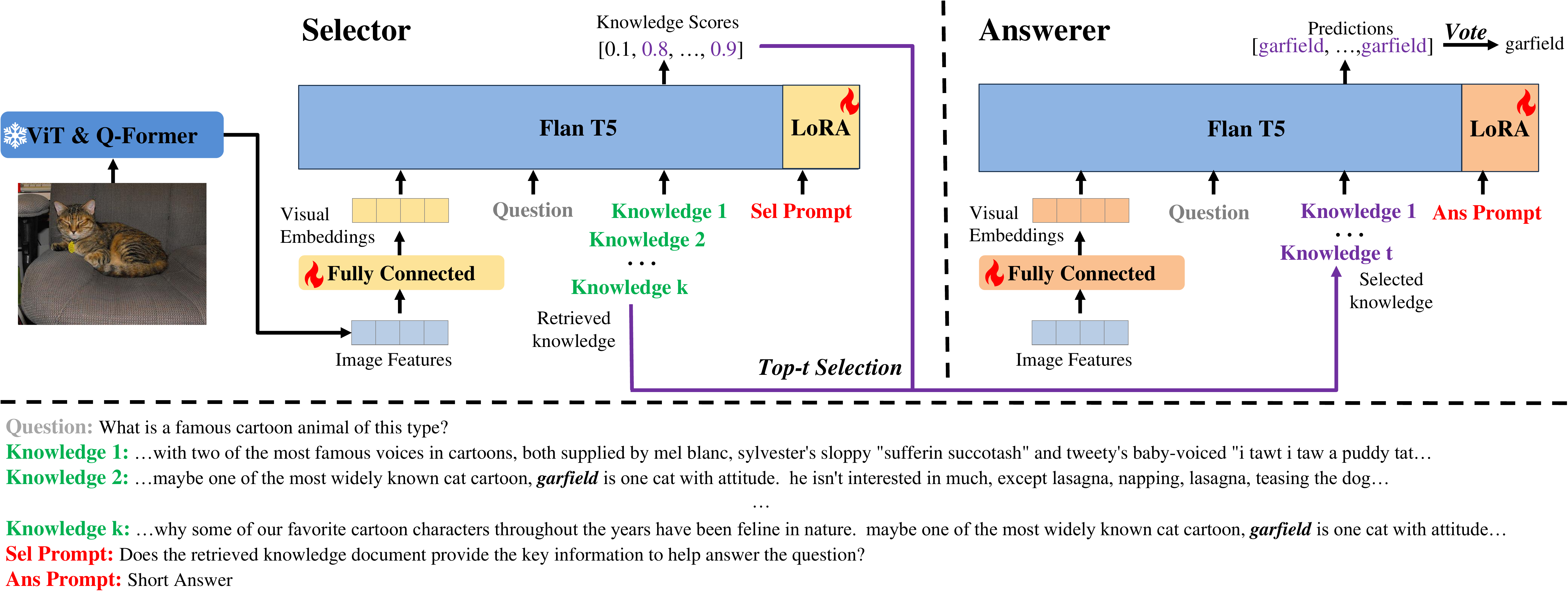}
   \caption{Our framework consists of two modules: a Selector and an Answerer.  Selector (left) selects the top-T knowledge documents for the Answerer (right), and the Answerer focuses on important knowledge information to predict answers. Both modules utilize the same frozen visual module to extract image features. We train the fully connected (FC) layer and fine-tune the language model using LoRA, which amounts to only 0.16\% of the total parameters. For detailed training procedures of the two modules, refer to Alg.~\ref{alg:alg1}. The original knowledge is retrieved using DPR, and for brevity, we omit the retrieval process here (details can be found in Section \ref{pre}).}
   \label{fig:framework}
\end{figure*}

\paragraph{Knowledge-based VQA.} Conventional VQA benchmarks \cite{goyal2017making, hudson2019gqa} primarily focus on basic visual perception and reasoning tasks and numerous studies have achieved promising results on these benchmarks \cite{Anderson2017BottomUpAT,zhang2021vinvl,tan2019lxmert, lu2019vilbert,li2022blip,wang2022ofa}. Different from them, the knowledge-based VQA task \cite{wang2017fvqa,marino2019ok,Schwenk2022AOKVQAAB} requires models to incorporate diverse world knowledge to respond to questions about visual content, which is more challenging.
Recent studies \cite{garderes2020conceptbert, wu2022multi, lin2022retrieval, gui2021kat} have explored various open-domain world knowledge sources, such as ConceptNet \cite{speer2017conceptnet}, Wikipedia \cite{vrandevcic2014wikidata}, Google Search Corpus \cite{luo2021weakly}. They retrieve the relevant knowledge documents from the knowledge bases and integrate them into the answering model to generate predictions. Except for using explicit knowledge, some methods also take GPT-3 \cite{brown2020language} as an implicit knowledge producer. They either prompt GPT-3 with in-context examples to predict answers directly \cite{yang2022empirical, hu2022promptcap, shao2023prompting}, or use GPT-3 to generate answer candidates with evidence serving as textual implicit knowledge bases \cite{gui2021kat, lin2022revive},  leading to significant performance improvements.
Different from these approaches, we employ a large visual-language model to select key retrieved knowledge and reason on the knowledge to answer questions. 

\section{Method}
In this section, we first introduce the preliminaries of Knowledge Retrieval and LVLM, which are the foundation of our framework. Then, we present the design of the Selector and Answerer for knowledge selection and question answering on knowledge respectively. Finally, we illustrate the self-bootstrap training method of two designed modules.
\subsection{Preliminaries}
\label{pre}
\paragraph{Knowledge Retrieval.} We adopt the Dense Passage Retrieval (DPR) \cite{karpukhin2020dense} to retrieve the knowledge documents. We transform the image into raw texts composed of captions, objects, attributes, and OCR (Optical Character Recognition). Then we compute the similarity scores between the query and knowledge documents $sim(q_i, D_j)=\mathbf{q}_i^{T}\cdot\mathbf{d}_j$ and exploit FAISS \cite{johnson2019billion} to index Top-k related knowledge documents $\mathcal{P}_i = \{P_{i,1}, P_{i,2}, ..., P_{i,k}\}$ for $i$-th query.


\paragraph{Large Visual-Language Model.} In our work, both knowledge selection and question-answering modules adopt BLIP-2 \cite{li2023blip} as the backbone. The architecture of BLIP-2 comprises a frozen image encoder \cite{dosovitskiy2020image, fang2023eva}, a Q-Former \cite{li2023blip}, and a pre-trained language model \cite{chung2022scaling}. Given an image $I_i$, the frozen image encoder outputs a set of visual features $\{\mathbf{h}_{i,1}, \mathbf{h}_{i,2}, ..., \mathbf{h}_{i,m}\}$.  Q-Former takes extracted visual features as input, and outputs language-aligned visual features $\{\mathbf{v}_{i,1}, \mathbf{v}_{i,2}, ..., \mathbf{v}_{i,l}\}$. These visual features are concatenated with the textual word embeddings, which are fed into the language model for generation. Through pre-training on large-scale image-caption datasets,  Q-Former can effectively project visual features into the feature space of the Language Large Model (LLM). We freeze the visual encoder and Q-former during training. We train the fully connected layer and use LoRA \cite{hu2022lora} to finetune the LLM (only finetune 0.16\% of total parameters). 


\subsection{Selector and Answerer}



\paragraph{Selector.} After obtaining the Top-k knowledge documents using DPR for the $i$-th sample, we aim to choose $t$ most important knowledge documents from the retrieved documents, where $t$ is smaller than $k$. As shown in Fig.~\ref{fig:framework}, we first use the frozen image encoder and Q-former to extract the image features $\mathbf{V}_i$, where these features are extracted once and then used by the Selector and the Answerer. Then image features $\mathbf{V}_i$ are fed into the independent fully-connected layer to obtain the visual embeddings $\mathbf{E}^v_i$. We concatenate the question, a retrieved knowledge document, and the Selection prompt "Does the retrieved knowledge document provide the key information to help answer the question?" into one sentence $S$. Next, visual embeddings $\mathbf{E}^v_i$ and the text are concatenated and fed into the LLM (Flan-T5 \cite{chung2022scaling} is adopted in our work). Last, we use the probability of generating the word ‘yes’ as the score of each retrieved knowledge document $P_{i,j}$, denoted as $s_{i,j} = \bm{LLM}(concat(\mathbf{E}^v_i, S_i))$, and we select top-t documents $\hat{\mathcal{P}}_i=\{ \hat{P}_{i,1},\hat{P}_{i,2}, ...,\hat{P}_{i,t}\}$ based on the scores. The Selector can be conceptualized as follows:
\begin{equation}
  \hat{\mathcal{P}}_i= = Selector(I_i, Q_i, \mathcal{P}_i), |\hat{\mathcal{P}}_i|=t
  \label{eq:2}
\end{equation}

\paragraph{Answerer.} After obtaining the selected knowledge documents, we aim to reason on the knowledge to answer questions. As shown in Fig.~\ref{fig:framework}, we process the same image features to obtain the different visual embeddings $\mathbf{E}^v_i$ via the fully-connected layer of the Answerer. Next, we concatenate the question and the knowledge into one sentence $S'$ using the template "Question: \{\} Knowledge: \{\} Answer: ". We concatenate the visual embeddings and the text, which are fed into the LLM with different LoRA parameters to get the answer. The model outputs corresponding answers based on different documents. The Answerer can be conceptualized as follows:
\begin{equation}
  a_i = Answerer(I_i, Q_i, \hat{\mathcal{P}}_i)
  \label{eq:3}
\end{equation}

Then the final answer is based on the majority vote. We also tried different knowledge reasoning methods, such as concatenating (the results can be seen in the ablation study).

\begin{algorithm}[t]
\caption{Pipeline of cycle training}\label{alg:alg1}
\begin{algorithmic}
\STATE 
\STATE \textbf{Input:} \\
KB-VQA dataset $\mathcal{D}=\{I_i, Q_i, \mathcal{A}_i | i=1, 2, \dots, N\}$; 
\\Retrieved knowledge documents $\mathcal{P}_i = \{P_i^1, P_i^2, \dots, P_i^k\}$;
$I_i$, $Q_i$, $\mathcal{P}_i$, and $\mathcal{A}_i$ denote image, question, document set, and answer set of $i$-th sample
\STATE \textbf{Output:} Knowledge selection model $Selector$; Question answering model $Answerer$
\FOR{sample in $\mathcal{D}$}
\STATE \textbf{Stage 1:} 
\STATE 1: Using $Selector$ to select top-t documents $\mathcal{\hat{P}}_i$ from the retrieved knowledge documents $\mathcal{P}_i$  as Eq.~\ref{eq:2}
\STATE 2: Finetuning $Answerer$ on $\{I_i, Q_i, \mathcal{\hat{P}}_i, \mathcal{A}_i\}$ supervised by the ground-truth answer as Eq.~\ref{eq:loss_ans}.
\STATE \textbf{Stage 2:} 
\STATE 1: Using $Answerer$ to predict answers for retrieved knowledge documents $\mathcal{P}_i$ as Eq.~\ref{eq:3}
\STATE 2: Generating to pseudo labels $\{y_{i,j}\}$ for retrieved knowledge documents $\mathcal{P}_i$ as Eq.~\ref{eq:pesudo_label}
\STATE 3: Finetuning $Selector$ on $\{I_i, Q_i, \mathcal{P}_i, \{y_{i,j}\}\}$ supervised by the pseudo label as Eq.~\ref{eq:loss_sel}.
\ENDFOR
\end{algorithmic}
\label{alg1}
\end{algorithm}

\subsection{Self-Bootstrap Learning}
To enable the Selector and Answerer to select key knowledge and answer questions, we bootstrap them with each other in a style of cycle training. We repeat the following process for the given $i$-th sample $\{I_{i}, Q_{i}, \mathcal{P}_i, \mathcal{A}_i \}$ of the training dataset:

\paragraph{Answerer Training.} We use Eq.~\ref{eq:2} to get the selected knowledge documents $\hat{\mathcal{P}}_i$. The image $I_i$ is fed into the frozen ViT and Q-former to obtain the image features $\mathbf{V}_i$. We use the trainable $FC_{ans}$ layer to output the visual embeddings $\mathbf{E}^v_{ans,i}$.
We concatenate the visual embedding, the question $Q_{i}$ and each selected knowledge document $\hat{P}_{i,j}$ to construct $t$ triplets for the sample, where $j={1,2, \dots, t}$. Then we finetune the Answerer with LoRA under the supervision of the ground truth answer $\mathcal{A}_{i}$:
\begin{equation}
    \label{eq:loss_ans}
    \begin{aligned}
    & \mathbf{E}^v_{ans,i}  = FC_{ans}(\mathbf{V}_i), \\
    & L_{ans} = - \sum_{j=1}^{t} \log LLM_{ans}(a_i^*|\mathbf{E}^v_{ans,i},Q_{i},\hat{P}_{i}^{j}), 
    \end{aligned}
\end{equation}
where $a_i^*$ is the most frequent answer in the human-annotated answer set $\mathcal{A}_{i}$.

\paragraph{Selector Training.}
We first use Eq.~\ref{eq:3} to predict answers based on each retrieved knowledge document  $P_{i,j}$. Then we assign pseudo labels to the retrieved documents according to model predictions and weak supervision labels \cite{luo2021weakly, lin2022retrieval, Lin2023FinegrainedLM}. We use "yes" and "no" as pseudo labels, where label a document as positive knowledge if Answerer can output the correct answer using that document and the document contains any of the answers in $\mathcal{A}_{i}$.
\begin{equation}
\label{eq:pesudo_label}
\begin{aligned}
    y_{i,j} = 
    \begin{cases}
    \text{yes}, \quad \textit{if} \ \ a_i=a_i^* \wedge \\ \quad\quad P_{i,j} \ \ \text{contains an answer in} \ \ \mathcal{A}_{i}\\
    \text{no}, \quad else
    \end{cases}
\end{aligned}
\end{equation}

\begin{table*}[t]
  \centering
  \caption{\textbf{Performance comparison with state-of-the-art (SOTA) methods on the OK-VQA dataset.} Knowledge Sources: \textbf{C}onceptNet (C); \textbf{W}ikipedia (W); \textbf{G}oogle \textbf{S}earch (GS); \textbf{G}oogle \textbf{I}mages (GI).
  The best result in the table is bolded.
  The results show that our method achieves the state-of-the-art performance.}
  \resizebox{\linewidth}{!}{
  \begin{tabular}{l|c|c|c|c|c}
    \hline
    Models &  Large Models & $K_{train}$& $K_{test}$ &Knowledge Resource & Acc (\%) \\
    \hline
    BAN+AN \cite{marino2019ok} & - &-&-& W & 25.6\\
    ConceptBERT \cite{garderes2020conceptbert} & - &-&-& C & 33.7 \\
    KRISP \cite{marino2021krisp}  &-&-&-&C+W&38.4\\
    Visual Retriever-Reader \cite{luo2021weakly}  & - &100&100& GS&39.2\\
    MAVEx \cite{wu2022multi}  & - &-&-& W+C + GI & 39.4\\
    \hline
    PICa \cite{yang2022empirical}  & GPT-3 (175B)&-&-& GPT-3 & 48.0\\
    TRiG(Ensemble) \cite{gao2022transform} & T5-large (770M) &100&100& W & 50.5 \\
    KAT(Single) \cite{gui2021kat} & T5-large (770M) &40&40& W + GPT-3 & 53.1 \\
    KAT(Ensemble) \cite{gui2021kat} & T5-large (770M) &40&40& W + GPT-3 & 54.4 \\
    RA-VQA \cite{lin2022retrieval}  & T5-large (770M) &5&50& GS & 54.5\\
    REVIVE(Single) \cite{lin2022revive}  & T5-large (770M) &40&40& W+GPT-3 & 56.6\\
    REVIVE(Ensemble) \cite{lin2022revive}  & T5-large (770M) &40&40& W+GPT-3 & 58.0\\
    PromptCap \cite{hu2022promptcap}  & GPT-3 (175B) &-&-& GPT-3 & 60.4\\
    Prophet \cite{shao2023prompting} & GPT-3 (175B) &-&-& GPT-3+MCAN &61.1\\
    FillingGap \cite{wang2023filling}&GPT-3 (175B) &-&-& GPT-3&61.3\\
    SimpleBaseline \cite{xenos2023simple}&LLaMA 2 (13B)&-&-&LLaMA 2&61.2\\
    Cola-FT \cite{chen2024large}&FLAN-T5(11B)&-&-&BLIP+OFA&62.4\\
    \hline
    Flamingo \cite{alayrac2022flamingo}  & Flamingo (80B) &-&-& Pretrain&57.8\\
    InstructBLIP \cite{Dai2023InstructBLIPTG}&  InstructBLIP Vicuna (7B)&-&-& Pretrain &62.1\\
    Qwen-VL \cite{bai2023qwen}&Qwen-VL(Qwen-7B)&-&-&Pretrain&58.6\\
    MM-Reasoner \cite{khademi-etal-2023-mm}&Flamingo (80B)&-&-&GPT-4&60.8\\
    \hline
    BLIP2 (fine-tuned) \cite{li2023blip}  & BLIP2 T5-XL (3B) &-&-& Pretrain& 55.4\\
    RA-VQA-v2 \cite{Lin2023FinegrainedLM} & BLIP2 T5-XL (3B)&5&5& GS &62.1\\
    PreFLMR \cite{Lin_Mei_Chen_Byrne_2024}&BLIP2 T5-XL (3B)&5&5&GS&61.8\\
    \textbf{Ours}  &BLIP2 T5-XL (3B) &5&5& GS & \textbf{62.8} \\
    \hline
  \end{tabular}}
  \label{tab:sota}
\end{table*}

After obtaining the pseudo label of each retrieved knowledge document, we use the trainable $FC_{sel}$ layer to output the visual embeddings $\mathbf{E}^v_{sel,i}$.
we concatenate the visual embedding, the question $Q_{i}$ and each retrieved knowledge document $P_{i,j}$ to construct $k$ triplets for the sample, where $j={1,2, \dots, k}$. Then we finetune the Selector with LoRA under the supervision of pseudo labels:
\begin{equation}
    \label{eq:loss_sel}
    \begin{aligned}
    & \mathbf{E}^v_{sel,i}  = FC_{sel}(\mathbf{V}_i) , \\
    & L_{sel} = - \sum_{j=1}^{k} \log LLM_{sel}(y_{i,j}|\mathbf{E}^v_{sel,i},Q_{i},P_{i}^{j})
    \end{aligned}
\end{equation}

We provide the overall training pipeline in Alg.~\ref{alg:alg1}. Through continuous iteration, the Selector will provide more crucial knowledge for the Answerer to accurately respond to questions. Meanwhile, the improvement in the Answerer's reasoning ability will also result in more precise pseudo-labeling, further enhancing the Selector's discriminative power. During the inference stage, we utilize the Selector to choose key knowledge, and then instruct the Answerer to respond to questions based on this knowledge.

\section{Experiments}
\subsection{Experimental Setup} 
\paragraph{Dataset.} We conduct extensive experiments on OK-VQA \cite{marino2019ok} to evaluate the effectiveness of our method. OK-VQA is a challenging open-domain knowledge-based VQA dataset that requires models to leverage various external knowledge sources to answer questions. The dataset contains 14,055 questions and 14,031 images, whereas the training set and testing set have 9k and 5k image-question pairs, respectively. Due to no knowledge base being provided for OK-VQA, we need to choose the proper knowledge base for the dataset. In this paper, we adopt Google Search Corpus \cite{luo2021weakly} as the knowledge base which is collected in the websites using the Google Search API. 

\paragraph{Evaluation Metric.} We use the standard VQA metric \cite{antol2015vqa} to evaluate the performance of the model. Given the prediction of the question $a$ and the groudtruth answer set $\mathcal{A}$, the VQA accuracy is calculated as:
\begin{equation}
    \label{eq:vqa_score}
    Accuracy(a,\mathcal{A}) = \min (\frac{\#A(a)}{3}, 1),
\end{equation}
where the groudtruth answer set $\mathcal{A}$ is annotated by different humans, $\#A(a)$ denotes the  occurrence of $a$ in $\mathcal{A}$.

\paragraph{Implementation Details.} In our experiment, we adopt BLIP2 T5-XL (3B) \cite{li2023blip} to initialize the Selector and Answerer. We freeze the image encoder and Q-former, with both the Selector and Answerer sharing the same visual module. We finetune the fully connected layer and use LoRA \cite{hu2022lora} to train the LLM. We use the default huggingface-PEFT setting: r=8, lora\_alpha=32, lora\_dropout=0.1. We use Adam as the optimizer and set the batch size to 8. We use the warm-up strategy which trains the model with an initial learning rate of 1e-4 and warm-up factor of 0.05 for 1000 steps and then utilizes a cosine annealing learning strategy with an initial learning rate of 1e-4 and a final learning rate of 0 after 10 epochs. 
We use top-30 knowledge documents retrieved by a pre-trained DPR \cite{lin2022retrieval} as candidates for Selector and use the selected top-5 documents from the 30 documents for the Answerer to train and infer, denoted as $K_{candidate}=30, K_{train}=5, K_{test}=5$. We use 2 Nvidia A800 GPUs (80G) for all experiments.

\subsection{Comparison with State-of-the-art Methods} As shown in Tab.~\ref{tab:sota}, we can see that early models (BAN+AN \cite{marino2019ok}, ConceptBERT \cite{garderes2020conceptbert}, KRISP \cite{marino2021krisp}, Visual Retriever-Reader \cite{luo2021weakly}, and MAVEx \cite{wu2022multi}) have a weak performance, achieving a VQA accuracy from 25.6\% to 39.4\%. Recently, by introducing larger models (T5-large, GPT-3, LLaMA, Vicuna) and diverse knowledge resources (ConceptNet, Wikipedia, Google Web Search and Google Images), the performance has a significant performance improvement, achieving a VQA accuracy of 62.4\%. Our method aims to augment the reasoning ability to answer knowledge-intensive questions of the large visual-language model. When directly finetuning BLIP2 T5-XL on OKVQA, the model has a low performance of 55.44\%. By introducing external knowledge, the performance has a significant performance improvement. Different from RA-VQA-v2 \cite{Lin2023FinegrainedLM} and PreFLMR \cite{Lin_Mei_Chen_Byrne_2024}, we do not train a multimodal retriever from scratch which requires expensive annotations and high computational costs. We directly leverage the large visual-language model to select key knowledge from the retrieved knowledge by DPR like the process of re-ranking. With the same knowledge resources (\emph{i.e.,} Google Search), our method achieves 62.8\% accuracy, outperforming other state-of-the-art models. It is worth noting that we do not use GPT-3 and we only train the 0.16\% parameters of the model. These results demonstrate the effectiveness of the proposed approach.

\subsection{Ablation Study}
We conduct the ablation studies to evaluate different components of our framework on OK-VQA.

\paragraph{Effect of Selector.} We conduct the ablation study to evaluate the effectiveness of Selector in our method. We show the results in Tab.~\ref{tab:ab_sel}. From the results, we can observe: our framework, leveraging key knowledge documents selected by the Selector, consistently outperforms the Answerer when using the same number of documents retrieved by DPR. We improve the performance by 2.14\% and 1.88\% with 1 and 5 test knowledge documents, compared to DPR-based retrieval. When using the randomly selected documents, the model performs worst. These results demonstrate that top-ranked knowledge documents based on DPR scores are not optimal for question answering and our key knowledge selection module can identify relevant documents for accurate question answering, ensuring the coherence of knowledge retrieval and question-answering processes. 
\begin{table}
  \caption{Comparison of our selector with different knowledge selection strategies. We select 5 knowledge documents from top-30 knowledge candidates retrieved by DPR. \textbf{DPR Score} refers to selecting top-5 knowledge based on similarity scores.  \textbf{Random Selection} means randomly selecting 5 knowledge documents from 30 candidate knowledge documents. \textbf{Selector} denotes choosing 5 key knowledge documents by the Selector.}
  \label{tab:ab_sel}
  \resizebox{\linewidth}{!}{
  \begin{tabular}{cc|c|c}
    \hline
    $K_{train}$&$K_{test}$&Knowledge Selection&Acc (\%) \\
    \hline
    5&1& Random Selection & 50.45\\
    5&1& DPR Score & 58.80\\
    5&1& Selector & 61.62\\
    \hline
    5&5& Random Selection & 55.05\\
    5&5& DPR Score& 60.69\\
    5&5& Selector & 62.83\\
  \bottomrule
\end{tabular}}
\end{table}

\paragraph{Effect of different knowledge reasoning methods of Answerer.} In Tab.~\ref{tab:ab_ans}, we present a comparison of Answerer using different knowledge reasoning methods. The results show that the performance using the strategy of voting surpasses that of concatenating under different knowledge selection settings. We argue that directly combining all the knowledge documents into a lengthened document makes it difficult for Answerer to reason on them, which is easily influenced by noisy information. In contrast, it is easier for Answerer to reason on each document to predict the answer. Simple voting can choose the best answer.
\begin{table}
  \caption{Effect of different knowledge reasoning methods of Answerer. \textbf{Concatenating} denotes that we combine the key knowledge documents into one sentence and feed it into Answerer to predict the final answer. \textbf{Voting} means that we feed different key knowledge documents into Answerer to predict different answers and choose the best answer based on majority voting.}
  \label{tab:ab_ans}
  \begin{tabular}{c|c|c}
    \hline
    Method&VQA Model&Acc (\%) \\
    \hline
    Concatenating&\multirow{2}{*}{BLIP2 (fine-tuned)}& 59.11\\
    Voting&  & 60.69\\
    \hline
    Concatenating&\multirow{2}{*}{Ours}& 62.06\\
    Voting&& 62.83\\
    
  \bottomrule
\end{tabular}
\end{table}

\paragraph{Effect of the self-bootstrap learning method.} To evaluate the effectiveness of our self-bootstrap learning method, we compare the method with the strategy of independent training of two modules. We finetune the Answerer with the knowledge documents retrieved by DPR as the \textbf{baseline}. \textbf{Independent training} means that we take two passes in answerer training and one pass for selector training on the entire dataset. \textbf{Cycle training} means that we train the answerer and selector on each batch data of the dataset simultaneously.
The results in Tab.~\ref{tab:ab_cycle} show that the model with cycle training outperforms the model with independent training by 3.81\%. The VQA score of using independent training is even lower than the baseline. These results demonstrate that our cycle training method can effectively boost the Selector and Answerer each other, which makes the model find key knowledge documents and leverage the knowledge to answer questions.
\begin{table}
\centering
  \caption{Effect of the self-bootstrap learning method.}
  \label{tab:ab_cycle}
  \begin{tabular}{c|c}
    \hline
    Method&Acc (\%) \\
    \hline
    Baseline & 60.69 \\
    Independent training & 59.02\\
    Cycle training & 62.83\\
  \hline
\end{tabular}
\end{table}


\paragraph{Effect of different methods of pseudo-labeling.} In Tab.~\ref{tab:ab_label}, we compare the model performance with different methods of pseudo-labeling. When using the model predictions as guidance, the model has a VQA score of 62.31\%. When adding the weak supervision as the guidance, the model's VQA score increases from 62.31\% to 62.83\%. The results demonstrate that using weak supervision labels preserves potentially useful documents, aiding the Answerer in accurately answering questions.
\begin{table}
\centering
  \caption{Ablation study on different methods of pseudo-labeling.}
  \label{tab:ab_label}
  \begin{tabular}{c|c|c}
    \hline
    \makecell{Model \\predictions}& \makecell{Weak supervision \\labels}&Acc (\%) \\
    \hline
    $\checkmark$ & & 62.31\\
    $\checkmark$ & $\checkmark$& 62.83\\
  \hline
\end{tabular}
\end{table}

\paragraph{Effect of key knowledge documents selection ranges and quantities.} In Tab.~\ref{tab:ab_num}, we evaluate key knowledge document selection using various numbers of candidate documents and selected documents. From the results, we have the following findings:
(1) As the number of selected documents increases, the model’s performance improves. This indicates that using more documents to train and test contributes to answering questions. 
(2) Using more documents for training can improve the performance a lot (the 2nd line $v.s.$ the last line). However, using more documents for testing has almost no improvement (the 3rd line $v.s.$ 4th line).
(3) When the number of candidate documents increases, the model’s performance improves. The result demonstrates that low-ranked documents based on DPR scores may contain useful information for question answering. It is necessary for the model to select key knowledge documents.
\begin{table}
\centering
  \caption{Ablation study on different numbers of candidate documents and selected documents.}
  \label{tab:ab_num}
  \begin{tabular}{c|c|c|c}
    \hline
    $K_{candidate}$ &$K_{train}$&$K_{test}$& Acc (\%) \\
    \hline
    5  &1&1& 57.90\\
    5  &1&5& 58.32\\
    10  &1&1& 58.61\\
    10  &1&5& 59.40\\
    \hline
    10  &5&5& 61.86\\
    15  &5&5& 62.31\\
    30  &5&5& 62.83\\
    30  &5&1& 61.62\\
  \hline
\end{tabular}
\end{table}

    

\begin{table}
\centering
  \caption{Ablation study on different documents selection in Answerer fine-tuning.}
  \label{tab:ab_difsel}
  \begin{tabular}{c|c|c}
    \hline
    \multicolumn{2}{c|}{Knowledge Selection} & \multirow{2}{*}{Acc (\%)} \\
    \cmidrule(r){1-2}
    Training &  Inference & \\
    \hline
    DPR & Selector & 62.31\\
    Selector & DPR & 60.75\\
    Selector & Selector & 62.83\\
    
  \hline
\end{tabular}
\end{table}

     
\begin{figure*}[t]
  \centering
   \includegraphics[width=0.95\linewidth]{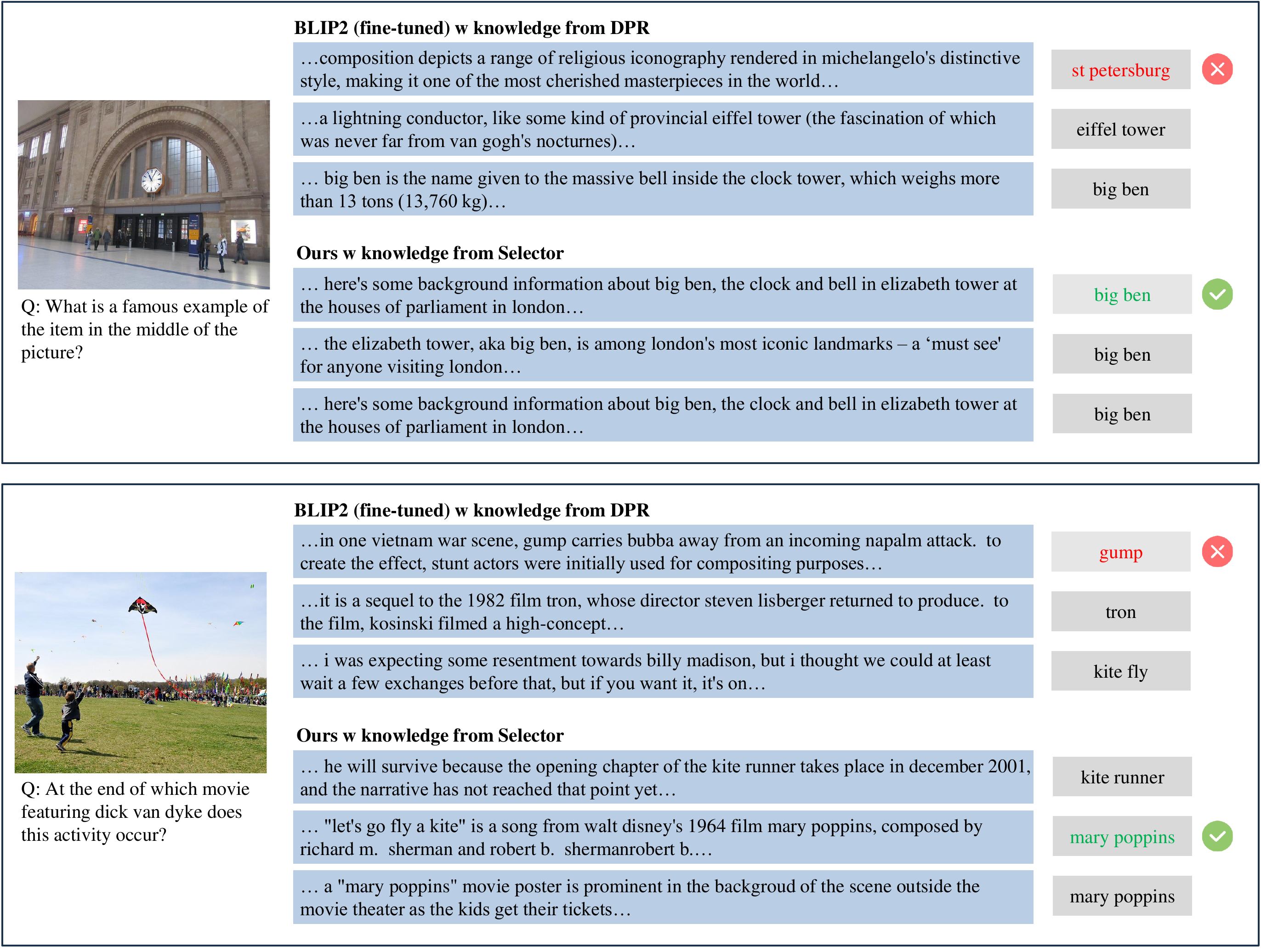}
   \caption{Qualitative results on the test split of OK-VQA. We compared our method with a model that fine-tunes BLIP2 with knowledge ranked by DPR. The middle segment of the graph represents knowledge from various methods used to answer questions. On the right side of the graph, different answers are depicted when using distinct knowledge. Green and red colors indicate whether the selected final answer is correct.}
   \label{fig:case}
\end{figure*}

\paragraph{Effect of different knowledge documents selection in Answerer fine-tuning.} Tab.~\ref{tab:ab_difsel} compares Answerer fine-tuning with different document selection strategies. The results show that our framework performs optimally when utilizing Selector in both Answerer training and inference. This is likely because the Selector provides more informative key knowledge documents and using both Selector ensures the consistency between the training domain and testing domain.

    
\paragraph{Performance of the knowledge retrieval.} In tab.~\ref{tab:retrieve}, we evaluate our Selector in the knowledge retrieval task. Following previous methods \cite{luo2021weakly,lin2022retrieval}, we adopt pseudo relevance to measure if the retrieved document is relevant to the query due to the absence of ground-truth document. We use Recall to measure the performance of the the knowledge retrieval. From the results, we can see that our Selector improves the performance of DPR a lot. This means our Selector can retrieve more relevant knowledge documents which help answer questions. Compared to other retrievers, our Selector achieves the second best performance. Although FLMR outperforms our Selector in the knowledge retrieval, our framework achieves better accuracy in VQA (shown in Tab.~\ref{tab:sota}) with the same backbone. This indicates that the knowledge documents selected by Selector have better consistency with Answerer.

\begin{table}
\centering
  \caption{Retrieval performance on Google Search (GS).}
  \label{tab:retrieve}
  \resizebox{\linewidth}{!}{
  \begin{tabular}{c|c|c}
    \hline
    Retriever & R@5 & R@10\\
    \hline
    VRR \cite{luo2021weakly}&80.4&88.55\\
    RA-VQA-FrDPR \cite{lin2022retrieval}&81.25&88.51\\
    RA-VQA \cite{lin2022retrieval}&82.84&89.00\\
    FLMR \cite{Lin2023FinegrainedLM} &\textbf{89.32}&\textbf{94.02}\\
    \hline
    DPR \cite{lin2022retrieval}&82.93&89.95\\
    \textbf{Our Selector} &88.66&93.56\\
  \hline
\end{tabular}}
\end{table}

\subsection{Qualitative Analysis}
In Fig.~\ref{fig:case}, We present a case study comparing our method with a model that fine-tunes BLIP2 using knowledge ranked by DPR. In the first case, top-ranked knowledge documents from DPR misguide the model, resulting in incorrect predictions. However, our method's Selector chooses key knowledge documents that aid in predicting correct answers. In the second case, each knowledge document from DPR contains irrelevant information, leading to an incorrect final answer. Despite the top-1 document from the Selector resulting in a wrong answer, our method identifies other key knowledge documents for generating correct answers. Through majority voting, the final selected answer is correct. These cases demonstrate our method's ability to extract informative knowledge from retrieved documents to support accurate question answering.

\section{Conclusion}
In this paper, we propose a novel framework that leverages the large visual-language model to construct two modules: (1) Selector for finding key retrieved knowledge and (2) Answerer for reasoning on the knowledge to predict answers.
We design a self-bootstrap learning method to improve their abilities, where the Selector chooses key knowledge documents for the Answerer and the Answerer provides pseudo-labels for the Selector.
Compared with state-of-the-art methods, our method achieves better performance on a challenging open-domain knowledge-based VQA benchmark (OK-VQA) and we conduct a comprehensive analysis to evaluate the effectiveness of our method.

\section{Acknowledgements}
This work was supported by the Artificial Intelligence-National Science and Technology Major Project (2023ZD0121200), the National Natural Science Foundation of China (U21B2043, 62102416), and the Key Research and Development Program of Jiangsu Province under Grant BE2023016-3. 

\section{Limitations}
Although our framework can effectively select key knowledge documents for answering question, it is inevitable that the knowledge still contains noise. In some cases, the model itself can answer the question without external knowledge, introducing extra knowledge may affect the performance. In the future, we can explore to dynamically select required knowledge to help itself answer questions.

In addition, there is a concern on the generalizability of the proposed method on other domains, especially when the initial DPR model can not retrieve gold standard context. In the future, we consider adopting a stronger multimodal retriever model to obtain more useful candidate knowledge documents, which enhances the  generalizability of our framework.
\bibliography{custom}

\begin{thebibliography}{52}
\providecommand{\natexlab}[1]{#1}

\bibitem[{vic(2023)}]{vicuna}
 2023.
\newblock Vicuna.
\newblock \url{https://github.com/lm-sys/FastChat}.

\bibitem[{Alayrac et~al.(2022)Alayrac, Donahue, Luc, Miech, Barr, Hasson, Lenc, Mensch, Millican, Reynolds et~al.}]{alayrac2022flamingo}
Jean-Baptiste Alayrac, Jeff Donahue, Pauline Luc, Antoine Miech, Iain Barr, Yana Hasson, Karel Lenc, Arthur Mensch, Katherine Millican, Malcolm Reynolds, et~al. 2022.
\newblock Flamingo: a visual language model for few-shot learning.
\newblock \emph{Advances in Neural Information Processing Systems}, 35:23716--23736.

\bibitem[{Anderson et~al.(2017)Anderson, He, Buehler, Teney, Johnson, Gould, and Zhang}]{Anderson2017BottomUpAT}
Peter Anderson, Xiaodong He, Chris Buehler, Damien Teney, Mark Johnson, Stephen Gould, and Lei Zhang. 2017.
\newblock \href {https://api.semanticscholar.org/CorpusID:3753452} {Bottom-up and top-down attention for image captioning and visual question answering}.
\newblock \emph{2018 IEEE/CVF Conference on Computer Vision and Pattern Recognition}, pages 6077--6086.

\bibitem[{Antol et~al.(2015)Antol, Agrawal, Lu, Mitchell, Batra, Zitnick, and Parikh}]{antol2015vqa}
Stanislaw Antol, Aishwarya Agrawal, Jiasen Lu, Margaret Mitchell, Dhruv Batra, C~Lawrence Zitnick, and Devi Parikh. 2015.
\newblock Vqa: Visual question answering.
\newblock In \emph{Proceedings of the IEEE international conference on computer vision}, pages 2425--2433.

\bibitem[{Bai et~al.(2023)Bai, Bai, Yang, Wang, Tan, Wang, Lin, Zhou, and Zhou}]{bai2023qwen}
Jinze Bai, Shuai Bai, Shusheng Yang, Shijie Wang, Sinan Tan, Peng Wang, Junyang Lin, Chang Zhou, and Jingren Zhou. 2023.
\newblock Qwen-vl: A frontier large vision-language model with versatile abilities.
\newblock \emph{arXiv preprint arXiv:2308.12966}.

\bibitem[{Brown et~al.(2020)Brown, Mann, Ryder, Subbiah, Kaplan, Dhariwal, Neelakantan, Shyam, Sastry, Askell et~al.}]{brown2020language}
Tom Brown, Benjamin Mann, Nick Ryder, Melanie Subbiah, Jared~D Kaplan, Prafulla Dhariwal, Arvind Neelakantan, Pranav Shyam, Girish Sastry, Amanda Askell, et~al. 2020.
\newblock Language models are few-shot learners.
\newblock \emph{Advances in neural information processing systems}, 33:1877--1901.

\bibitem[{Chen et~al.(2024)Chen, Li, Shen, Yang, Li, Keutzer, Darrell, and Liu}]{chen2024large}
Liangyu Chen, Bo~Li, Sheng Shen, Jingkang Yang, Chunyuan Li, Kurt Keutzer, Trevor Darrell, and Ziwei Liu. 2024.
\newblock Large language models are visual reasoning coordinators.
\newblock \emph{Advances in Neural Information Processing Systems}, 36.

\bibitem[{Chen et~al.(2021)Chen, Chen, Geng, Pan, Yuan, and Chen}]{chen2021zero}
Zhuo Chen, Jiaoyan Chen, Yuxia Geng, Jeff~Z Pan, Zonggang Yuan, and Huajun Chen. 2021.
\newblock Zero-shot visual question answering using knowledge graph.
\newblock In \emph{The Semantic Web--ISWC 2021: 20th International Semantic Web Conference, ISWC 2021, Virtual Event, October 24--28, 2021, Proceedings 20}, pages 146--162. Springer.

\bibitem[{Chung et~al.(2022)Chung, Hou, Longpre, Zoph, Tay, Fedus, Li, Wang, Dehghani, Brahma et~al.}]{chung2022scaling}
Hyung~Won Chung, Le~Hou, Shayne Longpre, Barret Zoph, Yi~Tay, William Fedus, Yunxuan Li, Xuezhi Wang, Mostafa Dehghani, Siddhartha Brahma, et~al. 2022.
\newblock Scaling instruction-finetuned language models.
\newblock \emph{arXiv preprint arXiv:2210.11416}.

\bibitem[{Dai et~al.(2023)Dai, Li, Li, Tiong, Zhao, Wang, Li, Fung, and Hoi}]{Dai2023InstructBLIPTG}
Wenliang Dai, Junnan Li, Dongxu Li, Anthony Meng~Huat Tiong, Junqi Zhao, Weisheng Wang, Boyang~Albert Li, Pascale Fung, and Steven C.~H. Hoi. 2023.
\newblock \href {https://api.semanticscholar.org/CorpusID:258615266} {Instructblip: Towards general-purpose vision-language models with instruction tuning}.
\newblock \emph{ArXiv}, abs/2305.06500.

\bibitem[{Dosovitskiy et~al.(2020)Dosovitskiy, Beyer, Kolesnikov, Weissenborn, Zhai, Unterthiner, Dehghani, Minderer, Heigold, Gelly et~al.}]{dosovitskiy2020image}
Alexey Dosovitskiy, Lucas Beyer, Alexander Kolesnikov, Dirk Weissenborn, Xiaohua Zhai, Thomas Unterthiner, Mostafa Dehghani, Matthias Minderer, Georg Heigold, Sylvain Gelly, et~al. 2020.
\newblock An image is worth 16x16 words: Transformers for image recognition at scale.
\newblock \emph{arXiv preprint arXiv:2010.11929}.

\bibitem[{Fang et~al.(2023)Fang, Wang, Xie, Sun, Wu, Wang, Huang, Wang, and Cao}]{fang2023eva}
Yuxin Fang, Wen Wang, Binhui Xie, Quan Sun, Ledell Wu, Xinggang Wang, Tiejun Huang, Xinlong Wang, and Yue Cao. 2023.
\newblock Eva: Exploring the limits of masked visual representation learning at scale.
\newblock In \emph{Proceedings of the IEEE/CVF Conference on Computer Vision and Pattern Recognition}, pages 19358--19369.

\bibitem[{Gao et~al.(2022)Gao, Ping, Thattai, Reganti, Wu, and Natarajan}]{gao2022transform}
Feng Gao, Qing Ping, Govind Thattai, Aishwarya Reganti, Ying~Nian Wu, and Prem Natarajan. 2022.
\newblock Transform-retrieve-generate: Natural language-centric outside-knowledge visual question answering.
\newblock In \emph{Proceedings of the IEEE/CVF Conference on Computer Vision and Pattern Recognition}, pages 5067--5077.

\bibitem[{Gard{\`e}res et~al.(2020)Gard{\`e}res, Ziaeefard, Abeloos, and Lecue}]{garderes2020conceptbert}
Fran{\c{c}}ois Gard{\`e}res, Maryam Ziaeefard, Baptiste Abeloos, and Freddy Lecue. 2020.
\newblock Conceptbert: Concept-aware representation for visual question answering.
\newblock In \emph{Findings of the Association for Computational Linguistics: EMNLP 2020}, pages 489--498.

\bibitem[{Goyal et~al.(2017)Goyal, Khot, Summers-Stay, Batra, and Parikh}]{goyal2017making}
Yash Goyal, Tejas Khot, Douglas Summers-Stay, Dhruv Batra, and Devi Parikh. 2017.
\newblock Making the v in vqa matter: Elevating the role of image understanding in visual question answering.
\newblock In \emph{Proceedings of the IEEE conference on computer vision and pattern recognition}, pages 6904--6913.

\bibitem[{Gui et~al.(2021)Gui, Wang, Huang, Hauptmann, Bisk, and Gao}]{gui2021kat}
Liangke Gui, Borui Wang, Qiuyuan Huang, Alex Hauptmann, Yonatan Bisk, and Jianfeng Gao. 2021.
\newblock Kat: A knowledge augmented transformer for vision-and-language.
\newblock \emph{arXiv preprint arXiv:2112.08614}.

\bibitem[{Guo et~al.(2022)Guo, Nie, Wong, Liu, Cheng, and Kankanhalli}]{guo2022unified}
Yangyang Guo, Liqiang Nie, Yongkang Wong, Yibing Liu, Zhiyong Cheng, and Mohan Kankanhalli. 2022.
\newblock A unified end-to-end retriever-reader framework for knowledge-based vqa.
\newblock In \emph{Proceedings of the 30th ACM International Conference on Multimedia}, pages 2061--2069.

\bibitem[{Hu et~al.(2022{\natexlab{a}})Hu, Shen, Wallis, Allen-Zhu, Li, Wang, Wang, and Chen}]{hu2022lora}
Edward~J Hu, Yelong Shen, Phillip Wallis, Zeyuan Allen-Zhu, Yuanzhi Li, Shean Wang, Lu~Wang, and Weizhu Chen. 2022{\natexlab{a}}.
\newblock \href {https://openreview.net/forum?id=nZeVKeeFYf9} {Lo{RA}: Low-rank adaptation of large language models}.
\newblock In \emph{International Conference on Learning Representations}.

\bibitem[{Hu et~al.(2022{\natexlab{b}})Hu, Hua, Yang, Shi, Smith, and Luo}]{hu2022promptcap}
Yushi Hu, Hang Hua, Zhengyuan Yang, Weijia Shi, Noah~A Smith, and Jiebo Luo. 2022{\natexlab{b}}.
\newblock Promptcap: Prompt-guided task-aware image captioning.
\newblock \emph{arXiv preprint arXiv:2211.09699}.

\bibitem[{Hudson and Manning(2019)}]{hudson2019gqa}
Drew~A Hudson and Christopher~D Manning. 2019.
\newblock Gqa: A new dataset for real-world visual reasoning and compositional question answering.
\newblock In \emph{Proceedings of the IEEE/CVF conference on computer vision and pattern recognition}, pages 6700--6709.

\bibitem[{Jiang et~al.(2018)Jiang, Natarajan, Chen, Rohrbach, Batra, and Parikh}]{jiang2018pythia}
Yu~Jiang, Vivek Natarajan, Xinlei Chen, Marcus Rohrbach, Dhruv Batra, and Devi Parikh. 2018.
\newblock Pythia v0. 1: the winning entry to the vqa challenge 2018.
\newblock \emph{arXiv preprint arXiv:1807.09956}.

\bibitem[{Johnson et~al.(2019)Johnson, Douze, and J{\'e}gou}]{johnson2019billion}
Jeff Johnson, Matthijs Douze, and Herv{\'e} J{\'e}gou. 2019.
\newblock Billion-scale similarity search with gpus.
\newblock \emph{IEEE Transactions on Big Data}, 7(3):535--547.

\bibitem[{Kamath et~al.(2022)Kamath, Clark, Gupta, Kolve, Hoiem, and Kembhavi}]{kamath2022webly}
Amita Kamath, Christopher Clark, Tanmay Gupta, Eric Kolve, Derek Hoiem, and Aniruddha Kembhavi. 2022.
\newblock Webly supervised concept expansion for general purpose vision models.
\newblock In \emph{European Conference on Computer Vision}, pages 662--681. Springer.

\bibitem[{Karpukhin et~al.(2020{\natexlab{a}})Karpukhin, Oguz, Min, Lewis, Wu, Edunov, Chen, and Yih}]{karpukhin-etal-2020-dense}
Vladimir Karpukhin, Barlas Oguz, Sewon Min, Patrick Lewis, Ledell Wu, Sergey Edunov, Danqi Chen, and Wen-tau Yih. 2020{\natexlab{a}}.
\newblock \href {https://doi.org/10.18653/v1/2020.emnlp-main.550} {Dense passage retrieval for open-domain question answering}.
\newblock In \emph{Proceedings of the 2020 Conference on Empirical Methods in Natural Language Processing (EMNLP)}, pages 6769--6781, Online. Association for Computational Linguistics.

\bibitem[{Karpukhin et~al.(2020{\natexlab{b}})Karpukhin, O{\u{g}}uz, Min, Lewis, Wu, Edunov, Chen, and Yih}]{karpukhin2020dense}
Vladimir Karpukhin, Barlas O{\u{g}}uz, Sewon Min, Patrick Lewis, Ledell Wu, Sergey Edunov, Danqi Chen, and Wen-tau Yih. 2020{\natexlab{b}}.
\newblock Dense passage retrieval for open-domain question answering.
\newblock \emph{arXiv preprint arXiv:2004.04906}.

\bibitem[{Khademi et~al.(2023)Khademi, Yang, Frujeri, and Zhu}]{khademi-etal-2023-mm}
Mahmoud Khademi, Ziyi Yang, Felipe Frujeri, and Chenguang Zhu. 2023.
\newblock \href {https://doi.org/10.18653/v1/2023.findings-emnlp.437} {{MM}-reasoner: A multi-modal knowledge-aware framework for knowledge-based visual question answering}.
\newblock In \emph{Findings of the Association for Computational Linguistics: EMNLP 2023}, pages 6571--6581, Singapore. Association for Computational Linguistics.

\bibitem[{Li et~al.(2023)Li, Li, Savarese, and Hoi}]{li2023blip}
Junnan Li, Dongxu Li, Silvio Savarese, and Steven Hoi. 2023.
\newblock Blip-2: Bootstrapping language-image pre-training with frozen image encoders and large language models.
\newblock \emph{arXiv preprint arXiv:2301.12597}.

\bibitem[{Li et~al.(2022)Li, Li, Xiong, and Hoi}]{li2022blip}
Junnan Li, Dongxu Li, Caiming Xiong, and Steven Hoi. 2022.
\newblock Blip: Bootstrapping language-image pre-training for unified vision-language understanding and generation.
\newblock In \emph{International conference on machine learning}, pages 12888--12900. PMLR.

\bibitem[{Lin and Byrne(2022)}]{lin2022retrieval}
Weizhe Lin and Bill Byrne. 2022.
\newblock Retrieval augmented visual question answering with outside knowledge.
\newblock \emph{arXiv preprint arXiv:2210.03809}.

\bibitem[{Lin et~al.(2023)Lin, Chen, Mei, Coca, and Byrne}]{Lin2023FinegrainedLM}
Weizhe Lin, Jinghong Chen, Jingbiao Mei, Alexandru Coca, and Bill Byrne. 2023.
\newblock \href {https://api.semanticscholar.org/CorpusID:263310932} {Fine-grained late-interaction multi-modal retrieval for retrieval augmented visual question answering}.
\newblock \emph{ArXiv}, abs/2309.17133.

\bibitem[{Lin et~al.(2024)Lin, Mei, Chen, and Byrne}]{Lin_Mei_Chen_Byrne_2024}
Weizhe Lin, Jingbiao Mei, Jinghong Chen, and Bill Byrne. 2024.
\newblock \href {http://arxiv.org/abs/2402.08327} {Preflmr: Scaling up fine-grained late-interaction multi-modal retrievers}.
\newblock (arXiv:2402.08327).

\bibitem[{Lin et~al.(2022)Lin, Xie, Chen, Xu, Zhu, and Yuan}]{lin2022revive}
Yuanze Lin, Yujia Xie, Dongdong Chen, Yichong Xu, Chenguang Zhu, and Lu~Yuan. 2022.
\newblock Revive: Regional visual representation matters in knowledge-based visual question answering.
\newblock \emph{Advances in Neural Information Processing Systems}, 35:10560--10571.

\bibitem[{Liu et~al.(2023)Liu, Li, Wu, and Lee}]{liu2023visual}
Haotian Liu, Chunyuan Li, Qingyang Wu, and Yong~Jae Lee. 2023.
\newblock Visual instruction tuning.
\newblock \emph{arXiv preprint arXiv:2304.08485}.

\bibitem[{Lu et~al.(2019)Lu, Batra, Parikh, and Lee}]{lu2019vilbert}
Jiasen Lu, Dhruv Batra, Devi Parikh, and Stefan Lee. 2019.
\newblock Vilbert: Pretraining task-agnostic visiolinguistic representations for vision-and-language tasks.
\newblock \emph{Advances in neural information processing systems}, 32.

\bibitem[{Luo et~al.(2021)Luo, Zeng, Banerjee, and Baral}]{luo2021weakly}
Man Luo, Yankai Zeng, Pratyay Banerjee, and Chitta Baral. 2021.
\newblock Weakly-supervised visual-retriever-reader for knowledge-based question answering.
\newblock \emph{arXiv preprint arXiv:2109.04014}.

\bibitem[{Marino et~al.(2021)Marino, Chen, Parikh, Gupta, and Rohrbach}]{marino2021krisp}
Kenneth Marino, Xinlei Chen, Devi Parikh, Abhinav Gupta, and Marcus Rohrbach. 2021.
\newblock Krisp: Integrating implicit and symbolic knowledge for open-domain knowledge-based vqa.
\newblock In \emph{Proceedings of the IEEE/CVF Conference on Computer Vision and Pattern Recognition}, pages 14111--14121.

\bibitem[{Marino et~al.(2019)Marino, Rastegari, Farhadi, and Mottaghi}]{marino2019ok}
Kenneth Marino, Mohammad Rastegari, Ali Farhadi, and Roozbeh Mottaghi. 2019.
\newblock Ok-vqa: A visual question answering benchmark requiring external knowledge.
\newblock In \emph{Proceedings of the IEEE/cvf conference on computer vision and pattern recognition}, pages 3195--3204.

\bibitem[{Radford et~al.(2021)Radford, Kim, Hallacy, Ramesh, Goh, Agarwal, Sastry, Askell, Mishkin, Clark et~al.}]{radford2021learning}
Alec Radford, Jong~Wook Kim, Chris Hallacy, Aditya Ramesh, Gabriel Goh, Sandhini Agarwal, Girish Sastry, Amanda Askell, Pamela Mishkin, Jack Clark, et~al. 2021.
\newblock Learning transferable visual models from natural language supervision.
\newblock In \emph{International conference on machine learning}, pages 8748--8763. PMLR.

\bibitem[{Schwenk et~al.(2022)Schwenk, Khandelwal, Clark, Marino, and Mottaghi}]{Schwenk2022AOKVQAAB}
Dustin Schwenk, Apoorv Khandelwal, Christopher Clark, Kenneth Marino, and Roozbeh Mottaghi. 2022.
\newblock \href {https://api.semanticscholar.org/CorpusID:249375629} {A-okvqa: A benchmark for visual question answering using world knowledge}.
\newblock In \emph{European Conference on Computer Vision}.

\bibitem[{Shao et~al.(2023)Shao, Yu, Wang, and Yu}]{shao2023prompting}
Zhenwei Shao, Zhou Yu, Meng Wang, and Jun Yu. 2023.
\newblock Prompting large language models with answer heuristics for knowledge-based visual question answering.
\newblock In \emph{Proceedings of the IEEE/CVF Conference on Computer Vision and Pattern Recognition}, pages 14974--14983.

\bibitem[{Speer et~al.(2017)Speer, Chin, and Havasi}]{speer2017conceptnet}
Robyn Speer, Joshua Chin, and Catherine Havasi. 2017.
\newblock Conceptnet 5.5: An open multilingual graph of general knowledge.
\newblock In \emph{Proceedings of the AAAI conference on artificial intelligence}, volume~31.

\bibitem[{Tan and Bansal(2019)}]{tan2019lxmert}
Hao Tan and Mohit Bansal. 2019.
\newblock Lxmert: Learning cross-modality encoder representations from transformers.
\newblock \emph{arXiv preprint arXiv:1908.07490}.

\bibitem[{Touvron et~al.(2023)Touvron, Martin, Stone, Albert, Almahairi, Babaei, Bashlykov, Batra, Bhargava, Bhosale et~al.}]{touvron2023llama}
Hugo Touvron, Louis Martin, Kevin Stone, Peter Albert, Amjad Almahairi, Yasmine Babaei, Nikolay Bashlykov, Soumya Batra, Prajjwal Bhargava, Shruti Bhosale, et~al. 2023.
\newblock Llama 2: Open foundation and fine-tuned chat models.
\newblock \emph{arXiv preprint arXiv:2307.09288}.

\bibitem[{Vrande{\v{c}}i{\'c} and Kr{\"o}tzsch(2014)}]{vrandevcic2014wikidata}
Denny Vrande{\v{c}}i{\'c} and Markus Kr{\"o}tzsch. 2014.
\newblock Wikidata: a free collaborative knowledgebase.
\newblock \emph{Communications of the ACM}, 57(10):78--85.

\bibitem[{Wang et~al.(2017)Wang, Wu, Shen, Dick, and Van Den~Hengel}]{wang2017fvqa}
Peng Wang, Qi~Wu, Chunhua Shen, Anthony Dick, and Anton Van Den~Hengel. 2017.
\newblock Fvqa: Fact-based visual question answering.
\newblock \emph{IEEE transactions on pattern analysis and machine intelligence}, 40(10):2413--2427.

\bibitem[{Wang et~al.(2022)Wang, Yang, Men, Lin, Bai, Li, Ma, Zhou, Zhou, and Yang}]{wang2022ofa}
Peng Wang, An~Yang, Rui Men, Junyang Lin, Shuai Bai, Zhikang Li, Jianxin Ma, Chang Zhou, Jingren Zhou, and Hongxia Yang. 2022.
\newblock Ofa: Unifying architectures, tasks, and modalities through a simple sequence-to-sequence learning framework.
\newblock In \emph{International Conference on Machine Learning}, pages 23318--23340. PMLR.

\bibitem[{Wang et~al.(2023)Wang, Chen, Li, and Liu}]{wang2023filling}
Ziyue Wang, Chi Chen, Peng Li, and Yang Liu. 2023.
\newblock Filling the image information gap for vqa: Prompting large language models to proactively ask questions.
\newblock \emph{arXiv preprint arXiv:2311.11598}.

\bibitem[{Wu et~al.(2022)Wu, Lu, Sabharwal, and Mottaghi}]{wu2022multi}
Jialin Wu, Jiasen Lu, Ashish Sabharwal, and Roozbeh Mottaghi. 2022.
\newblock Multi-modal answer validation for knowledge-based vqa.
\newblock In \emph{Proceedings of the AAAI conference on artificial intelligence}, volume~36, pages 2712--2721.

\bibitem[{Xenos et~al.(2023)Xenos, Stafylakis, Patras, and Tzimiropoulos}]{xenos2023simple}
Alexandros Xenos, Themos Stafylakis, Ioannis Patras, and Georgios Tzimiropoulos. 2023.
\newblock A simple baseline for knowledge-based visual question answering.
\newblock In \emph{Proceedings of the 2023 Conference on Empirical Methods in Natural Language Processing}, pages 14871--14877.

\bibitem[{Yang et~al.(2022)Yang, Gan, Wang, Hu, Lu, Liu, and Wang}]{yang2022empirical}
Zhengyuan Yang, Zhe Gan, Jianfeng Wang, Xiaowei Hu, Yumao Lu, Zicheng Liu, and Lijuan Wang. 2022.
\newblock An empirical study of gpt-3 for few-shot knowledge-based vqa.
\newblock In \emph{Proceedings of the AAAI Conference on Artificial Intelligence}, volume~36, pages 3081--3089.

\bibitem[{Zhang et~al.(2021)Zhang, Li, Hu, Yang, Zhang, Wang, Choi, and Gao}]{zhang2021vinvl}
Pengchuan Zhang, Xiujun Li, Xiaowei Hu, Jianwei Yang, Lei Zhang, Lijuan Wang, Yejin Choi, and Jianfeng Gao. 2021.
\newblock Vinvl: Revisiting visual representations in vision-language models.
\newblock In \emph{Proceedings of the IEEE/CVF conference on computer vision and pattern recognition}, pages 5579--5588.

\bibitem[{Zhang et~al.(2022)Zhang, Roller, Goyal, Artetxe, Chen, Chen, Dewan, Diab, Li, Lin et~al.}]{zhang2022opt}
Susan Zhang, Stephen Roller, Naman Goyal, Mikel Artetxe, Moya Chen, Shuohui Chen, Christopher Dewan, Mona Diab, Xian Li, Xi~Victoria Lin, et~al. 2022.
\newblock Opt: Open pre-trained transformer language models.
\newblock \emph{arXiv preprint arXiv:2205.01068}.

\end{thebibliography}

\appendix

\section{Appendix}
\label{sec:appendix}

\begin{table}[t]
  \centering
  \caption{Performance comparison with state-of-the-art (SOTA) methods on the
FVQA dataset.}
  \resizebox{1\linewidth}{!}{
  \begin{tabular}{c|c}
    \toprule
    Method& Acc-1\\
    \hline
    Human&77.99\\
    UnifER \cite{guo2022unified}&55.04\\
    FVQA \cite{wang2017fvqa}&56.91\\
    ZS-VQA \cite{chen2021zero}&58.27\\
    FVQA(Ensemble) \cite{wang2017fvqa}&58.76\\
    MM-Reasoner(Ensemble) \cite{khademi-etal-2023-mm}&61.10\\
    \hline
    \textbf{Ours}& \textbf{63.3}\\
    \bottomrule
  \end{tabular}}
  \label{tab:fvqa}
\end{table}

\begin{table}[t]
  \centering
  \caption{Performance comparison with state-of-the-art (SOTA) methods on the
A-OKVQA dataset.}
  \resizebox{1\linewidth}{!}{
  \begin{tabular}{c|cc}
    \toprule
    \multirow{2}*{Method}& \multicolumn{2}{c}
    {Direct Answer}\\
    ~ &val& test\\
    \hline
    ClipCap \cite{Schwenk2022AOKVQAAB}  &18.1 &15.8 \\
    Pythia \cite{jiang2018pythia}  &25.2 &21.9 \\
    ViLBERT \cite{lu2019vilbert}&30.6 &25.9 \\
    LXMERT \cite{tan2019lxmert} &30.7 &25.9 \\
    KRISP \cite{marino2021krisp} &33.7 &27.1 \\
    GPV-2 \cite{kamath2022webly} &48.6 &40.7 \\
    BLIP-2 T5-XL \cite{li2023blip}&53.2 &49.7\\
    PromptCap + GPT-3 \cite{hu2022promptcap} &56.3 &\textbf{59.6} \\
    \hline
    \textbf{Ours} &\textbf{57.2} &56.4\\
    \bottomrule
  \end{tabular}}
  \label{tab:aok}
\end{table}

\subsection{Experiments on Other Datasets.} We also evaluate our method on FVQA \cite{fang2023eva} and A-OKVQA \cite{Schwenk2022AOKVQAAB} to demonstrate the effectiveness of our method. FVQA is a VQA dataset that mostly contains questions requiring external knowledge to answer, and provides supporting fact triplets alongside the image-question-answer triplets. A-OKVQA is an augmented successor of OK-VQA, containing 25K image-question pairs that require broader commonsense and world knowledge to answer. Due to A-OKVQA does not provide the knowledge source, we use Wikipedia \cite{vrandevcic2014wikidata} as the knowledge base.

As shown in Tab.~\ref{tab:fvqa}, our method surpasses previous state-of-the-art methods, which demonstrates the effectiveness and generalization of our method. Tab.~\ref{tab:aok} shows the comparative results on the challenging A-OKVQA dataset. Our method achieved competitive results, which demonstrates the effectiveness of our method.

\begin{table}[h]
\centering
  \caption{Computational cost of our framework.}
  \label{tab:cost}
  \begin{tabular}{c|c|c}
    \hline
    $K_{candidate}$ &Memory (GB)	&\makecell{Running Time \\ (sec./sample)}\\
    \hline
    10  &21.3	&1.0\\
    15	&21.4	&1.1\\
    30	&22.7	&1.3\\
  \hline
\end{tabular}
\end{table}

\subsection{Evaluation of Computational Cost}

In Tab.~\ref{tab:cost}, we show the computational cost of our framework using different knowledge candidates. As the number of candidate knowledge, the computational cost of our framework has a small increase. 

\end{document}